\documentclass[11pt]{article}
\usepackage[preprint]{acl}

\usepackage{times}
\usepackage{latexsym}
\usepackage{xurl}
\usepackage{hyperref}
\usepackage{tcolorbox}
\usepackage{geometry} 
\tcbuselibrary{skins, breakable,raster}
\usepackage[T1]{fontenc}

\usepackage[utf8]{inputenc}

\usepackage{microtype}

\usepackage{inconsolata}

\usepackage{graphicx}
\usepackage{xspace}
\usepackage{dirtytalk}
\usepackage[dvipsnames,table,xcdraw]{xcolor}

\usepackage{float}
\usepackage{multirow}
\usepackage{enumitem}
\usepackage{algorithm}
\usepackage{algpseudocode}
\usepackage{amsmath}
\usepackage{amsfonts}
\usepackage{amsthm}
\usepackage{amssymb}
\usepackage{cleveref}
\usepackage{booktabs}
\usepackage{siunitx} 
\newcommand{\directid}{\textsc{details}\@\xspace} 
\newcommand{\family}{\textsc{family}\@\xspace}  
\newcommand{\body}{\textsc{appearance}\@\xspace}   
\newcommand{\details}{\textsc{circumstances}\@\xspace}   
\newcommand{\socio}{\textsc{sec}\@\xspace} 
\newcommand{\fclt}{\textsc{fclt\_personnel}\@\xspace}   
\newcommand{\reltime}{\textsc{time}\@\xspace} 
\newcommand{\lfstl}{\textsc{lfstl}\@\xspace}  
\newcommand{\other}{\textsc{other}\@\xspace}   

\title{\textsc{DP-IPI}: A Hybrid Differential Privacy Text Rewriting Mechanism\\ for Indirect Personal Identifiers in Clinical Texts}

\author{
 \textbf{Ibrahim Baroud\textsuperscript{1,2}\thanks{These authors contributed equally.}}{\normalfont ,}
 \textbf{Stephen Meisenbacher\textsuperscript{3,4}\footnotemark[1]}{\normalfont ,}
 \\
 \textbf{Sebastian Möller\textsuperscript{1,2}},
 \textbf{Florian Matthes\textsuperscript{3}},
 \textbf{Roland Roller\textsuperscript{2}}
\\
 \textsuperscript{1}Quality \& Usability Lab, Technical University of Berlin, Berlin, Germany \\
 \textsuperscript{2}German Research Center for Artificial Intelligence (DFKI), Berlin, Germany \\
 \textsuperscript{3}Technical University of Munich, School of CIT, Garching, Germany \\
 \textsuperscript{4}Munich Center for Machine Learning, Munich, Germany
\\
 \small{
   \textbf{Correspondence:} \href{mailto:ibrahim.baroud@tu-berlin.de}{ibrahim.baroud@tu-berlin.de}, \href{mailto:stephen.meisenbacher@tum.de}{stephen.meisenbacher@tum.de}
 }
}

\newtcolorbox{examplebox}[2][]{%
  colback=orange!5!white, 
  colframe=gray!50!black, 
  coltitle=white, 
  title={#2},
  fonttitle=\bfseries,
  fontupper=\small,
  arc=1mm,
  #1
}

\begin{document}
\maketitle
\begin{abstract}
Despite the strengths of modern anonymization and de-identification techniques, the risk of re-identification remains significant due to the indirect identifiers remaining in texts. To address this problem, recent works have applied text rewriting under Differential Privacy (DP) to prevent data linkage by perturbing texts via noise addition. Such methods privatize \textit{all} tokens in a text indiscriminately, diminishing text quality and usability in critical domains such as in clinical settings. Focusing on \textit{indirect personal identifiers} (IPIs), we introduce a utility-preserving DP text rewriting method that only privatizes spans containing IPIs. We show that our method effectively reduces re-identification risks in clinical texts while being producing more coherent and usable output texts, leading to higher privacy-utility trade-offs. In this, we demonstrate the effectiveness of \textit{hybrid} text privatization, which leverages the promise of DP in an efficient, usable manner.
\end{abstract}

\section{Introduction}

\begin{figure*}[ht!]
    \centering
    \includegraphics[width=0.95\linewidth]{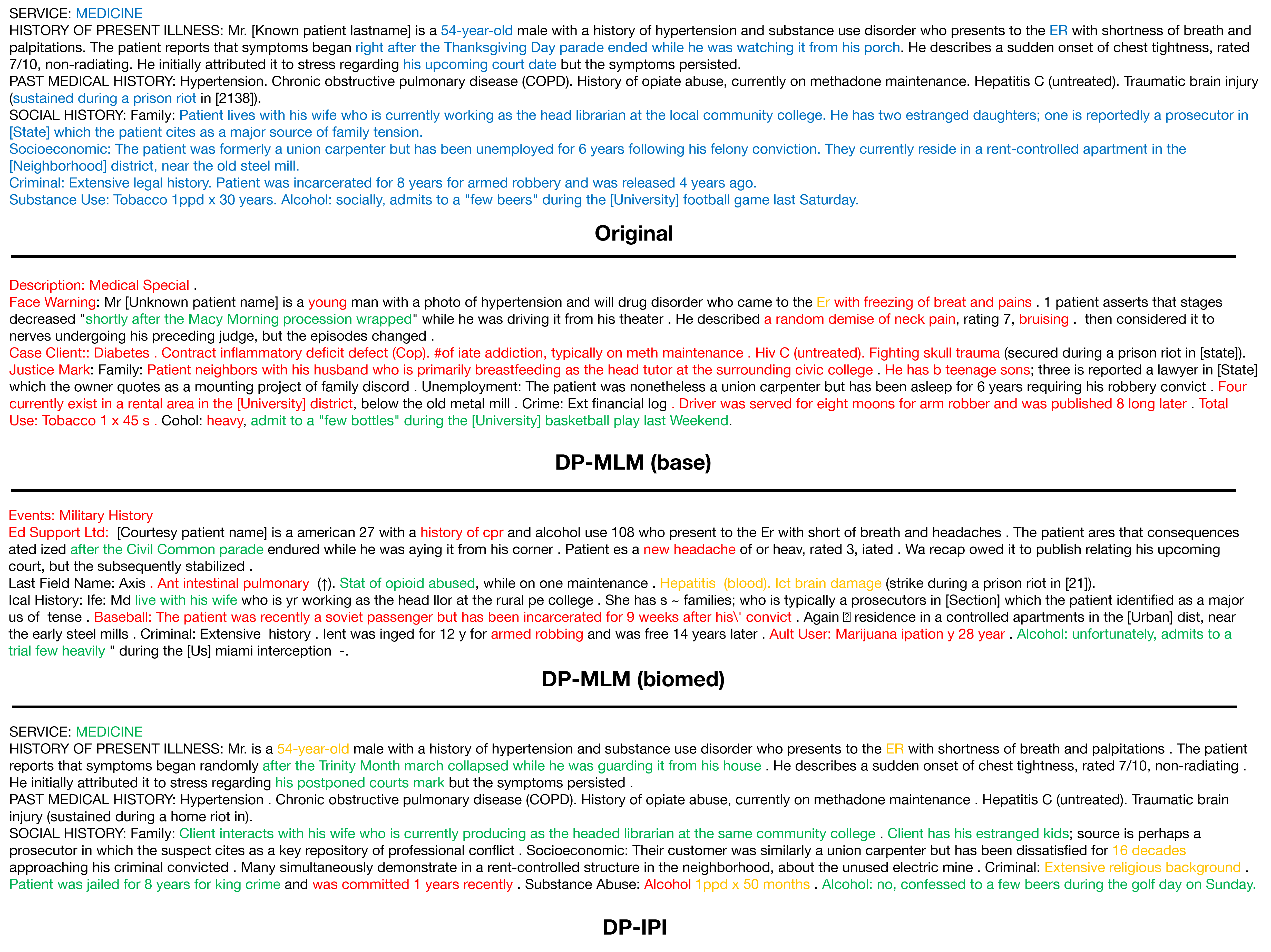}
    \caption{An illustrative example of our proposed \textsc{DP-IPI} compared to the original \textsc{DP-MLM} (at $\varepsilon=25$). In the original text, we highlight \textit{indirect personal identifiers} (IPIs) extracted by our trained IPI model (in \textcolor{RoyalBlue}{blue}). In the two corresponding privatized texts, we highlight aspects which are wrong, misleading, or hallucinated (\textcolor{red}{red}), borderline acceptable (\textcolor{YellowOrange}{orange}), and suitable IPI replacements (\textcolor{ForestGreen}{green}). These annotations also include medical aspects (e.g. allergies, medications) that are either improperly privatized (\textcolor{red}{red}) or correctly left unperturbed (\textcolor{ForestGreen}{green}). Note: the example is a synthetically generated patient file (using Google AI Studio), meant to mirror the MIMIC-III dataset.}
    \label{fig:dp-ipi}
\end{figure*}

In response to growing privacy concerns regarding the large-scale processing of text data, a plethora of research fields has developed in the realm of privacy-preserving Natural Language Processing \cite{sousa2023keep,YAN2025100300}, which specifically addresses the balance between data utility and  individual privacy. While the range of privacy-preserving NLP techniques is wide and varied, the ultimate challenge is two-fold: to determine what precisely in text is \say{private} \cite{10.1145/3531146.3534642}, and moreover, how best to privatize these aspects in text (while still maintaining downstream usability) \cite{meisenbacher2024comparative}.

Earlier streams of research focus on the task of \textit{anonymization} \cite{lison-etal-2021-anonymisation}, which largely emphasizes the identification, masking, and/or removal of direct identifiers in text, e.g. names or unique IDs. 
Removing direct identifiers is often not sufficient, since it leaves various information behind, such as profession, gender, or ethnicity, which might lead to re-identification. 
Therefore, studies have extended the detection of private information to include a variety of \textit{indirect} personal identifiers (IPIs) \cite{feder2020active,baroud-etal-2025-beyond}.
The treatment of both direct and indirect identifiers is especially important in scenarios where information cues can collectively re-identify individuals, such as with clinical notes \cite{sousa2023keep}.

Based on the framework of \textit{Differential Privacy} (DP) \cite{dwork2006differential}, recent state-of-the-art work focuses 
on how to integrate DP into NLP pipelines successfully, such that the required perturbations (i.e., via additive noise) can still produce meaningful and useful output data \cite{klymenko-etal-2022-differential}. 
DP-based text privatization methods can take many forms \cite{hu-etal-2024-differentially}, but specifically in text-in-text-out methods, the modus operandi involves perturbing units of text—words, sentences, or entire documents—under DP, and generating outputs that are ideally both privacy- and utility-preserving \cite{10.1145/3336191.3371856}. 
Despite theoretical privacy guarantees of such methods, recent literature points to limitations in coherent outputs and human acceptance \cite{igamberdiev-habernal-2023-dp,weiss-etal-2024-share}. These challenges render the practical usability of DP-based privatization methods low for critical domains, such as with clinical notes, where information fidelity matters greatly.

Despite continued progress, there remains a dearth of methods that explore the intersection of identifier-based anonymization and DP-based text privatization, and furthermore, studies therein that focus specifically on the intricacies of clinical text privatization.
The intersection of these two approaches create an valuable point of study for the biomedical domain, where privacy is paramount but unambiguous language is crucial. However, a hybrid method leveraging the strengths of both fields and focusing on usability for challenging domains has not been undertaken.

We introduce \textsc{DP-IPI}: a hybrid, efficient, and utility-preserving text privatization method that grounds itself in \textbf{DP}-based token privatization via the detection of \textit{\textbf{I}ndirect \textbf{P}ersonal \textbf{I}dentifiers} in clinical notes. 
This method improves previous DP mechanisms by perturbing tokens which constitute IPIs, thereby only privatizing the most sensitive tokens of a text. 
The focus on personal identifiers is motivated by traditional text anonymization, and by unifying this scope with the privacy guarantees offered by DP, we devise a method that is privacy-preserving where it matters for individual privacy protection, and utility-preserving where it does not.

Our evaluations center on privatizing clinical notes, a task not previously undertaken in the DP-based rewriting literature. 
We show that \textsc{DP-IPI} achieves near baseline utility while still minimizing re-identification risks, thereby achieving more favorable privacy-utility trade-offs than comparable methods which perturb complete texts. 
We thereby fuse identifier- and DP-based text privatization with the following contributions:

\begin{enumerate}[leftmargin=1.3em]
    \itemsep -0.3em
    \item We design an efficient text privatization mechanism which utilizes IPI detection to perform focused DP-based token perturbations.
    \item We evaluate \textsc{DP-IPI} on clinical notes against comparable methods and in corresponding ablations, leading to a critical discussion as well as clear recommendations for follow-up work.
    \item We release \textsc{DP-IPI} as an extension of \textsc{DP-MLM}: \url{https://github.com/sjmeis/DPMLM}.
\end{enumerate}

\section{Related Work}

\subsection{Differential Privacy in NLP}
Introduced by \citet{dwork2006differential}, Differential Privacy mathematically defines the notion of privacy in data sharing scenarios by bounding the influence a single individual's data can have on any computations performed on datasets. To realize DP, the de facto method has been the introduction of noise to either data points or the results of computations, so as to inject plausible deniability regarding the true values or attributes of the underlying individuals. While DP was originally envisioned for structured, relational databases (i.e., where each \say{individual} is a row of data), it has been extended and generalized to countless other data domains \cite{chatzikokolakis2013broadening,desfontaines2020sok}.

Recently, an increasing effort has been made to adapt the application of Differential Privacy to the domain of NLP at both model and data level. In applying DP to data, the notion of \textit{local} DP is often leveraged \cite{4690986}, which enables perturbation operations on single data points (i.e., texts). Approaches to achieving DP in text privatizations range from earlier word-level techniques \cite{10.1145/3336191.3371856, yue-etal-2021-differential} to more recent methods leveraging encoder-decoder \cite{bo-etal-2021-er,igamberdiev-habernal-2023-dp}, encoder-only \cite{meisenbacher-etal-2024-dp}, or decoder-only language models \cite{mattern-etal-2022-limits,utpala-etal-2023-locally,meisenbacher-etal-2025-impact}. The commonality between these methods is the transformation (i.e., privatization) of input units (words, tokens, etc.) to private counterparts; this is often framed as \textit{private text rewriting} \cite{igamberdiev-etal-2022-dp}.

Crucial to the success of DP text rewriting methods is the evaluation of \textit{effective} text privatization \cite{mattern-etal-2022-limits}. While a great deal of the evaluation of DP in NLP focuses on the privacy-utility trade-off and tuning of the privacy budget ($\varepsilon$) \cite{igamberdiev-habernal-2023-dp,meisenbacher2024comparative}, recent works have highlighted important factors with DP text privatization, such as group size \cite{vu-etal-2024-granularity}, optimal budget distribution within a text \cite{10.1145/3714393.3726504}, or ensuring proper empirical privacy protections \cite{tong-etal-2025-vulnerability}.
Importantly, studies have emphasized the challenge in producing DP outputs that are syntactically and structurally correct \cite{mattern-etal-2022-limits,arnold-2025-inspecting}. A less studied factor, however, is \textit{which} parts of a text are pertinent to privatize, and accordingly, how to leverage DP selectively on only the most privacy-sensitive spans. We address this gap through the lens of IPIs.

\subsection{Indirect Personal Identifiers (IPIs)}
Textual data, especially clinical texts, are rich with indirect information about individuals, which can be sufficient when combined to identify individuals \cite{sweeney2002k}. 
Unlike direct identifiers such as names and numerical identifiers, the definition and detection of IPIs is challenging due to their high linguistic complexity and infrequency \cite{baroud-etal-2025-beyond}. 
A limited number of works have studied indirect identifiers in textual data. 
\citet{feder2020active} defined such categories in the context of clinical texts as demographic traits and developed detection models on the sentence level. 
These categories included information such as marital status, occupation, and ethnicity. 
Moreover, beyond traditional de-identification, \citet{kolditz2019annotating} introduced the category of \textit{medical unit}, including information such as names of hospitals, hospital departments, and ambulant medical units.
\citet{baroud-etal-2025-beyond} expanded the categories from \citet{feder2020active} and  \citet{kolditz2019annotating} to cover a wider range of IPIs, e.g., circumstantial details, time expressions, and socio-economic and criminal history.

\section{Methodology}
Our \textsc{DP-IPI} method has two primary components: an IPI detection model and a DP-based rewriting model, introduced in the following. An illustrative example of \textsc{DP-IPI} can be found in Figure \ref{fig:dp-ipi}.

\subsection{IPI Detection} The first stage of \textsc{DP-IPI} is the detection of spans which carry indirect personal information. 
Although the definition and scope of IPI can be defined in various manners,  we adopt in our work the schema and annotations created by \citet{baroud-etal-2025-beyond}. 
In this work a dataset is introduced with around 6,200 annotations based on 100 discharge summaries from the Medical Information Mart for Intensive Care (MIMIC-III) \cite{johnson2016mimic} clinical dataset. 
The annotations cover the categories: \textsc{Appearance}, \textsc{Family}, \textsc{Circumstances}, \textsc{Healthcare Facilities and Personnel}, \textsc{Socioeconomic and Criminal History}, \textsc{Relative time}, \textsc{Details about a Direct Identifier}, \textsc{Hobbies and Lifestyle}, and \textsc{Other}. Further details about the definitions and characteristics of these categories and the annotation guidelines can be found in \citet{baroud-etal-2025-beyond}. 
Using the IPI annotations, we fine-tuned the \textsc{roberta-large} model by \citet{DBLP:journals/corr/abs-1907-11692} for token classification following \citet{otto-etal-2026-evaluating}. 
The fine-tuned IPI model can be found on Hugging Face,\footnote{\scriptsize\url{https://hf.co/Ibrahimbaroud/roberta-large-ipi-mimiciii}} and it is included as the default IPI model in the IPI integration of \textsc{DP-MLM}.
The performance on the held-out test set is provided in \Cref{tab:ipi_performance}.

\begin{table}[t]
\centering
\small
\begin{tabular}{l S S S S}
\toprule
\textbf{Category} & \textbf{P} & \textbf{R} & \textbf{F1} & \textbf{Support} \\
\midrule
\directid     & 0.08 & 0.25 & 0.12 & 4 \\
\family         & 0.67 & 1.00 & 0.80 & 73 \\
\body     & 0.61 & 0.69 & 0.65 & 29 \\
\details        & 0.20 & 0.37 & 0.26 & 30 \\
\socio          & 0.75 & 0.86 & 0.80 & 14 \\
\fclt   & 0.82 & 0.95 & 0.88 & 362 \\
\reltime & 0.88 & 0.97 & 0.92 & 1006 \\
\lfstl          & 0.65 & 0.94 & 0.77 & 35 \\
\other          & 0.50 & 0.71 & 0.59 & 7 \\
\midrule
Micro average  & 0.81 & 0.94 & 0.87 & 1560 \\
Macro average  & 0.57 & 0.75 & 0.64 & 1560 \\
\bottomrule
\end{tabular}
\caption{Evaluation results on the test set for \textsc{RoBERTa-large} in \textbf{P}recision, \textbf{R}ecall, and $\mathbf{F_1}$ score. Support shows the number of examples in the test set.}
\label{tab:ipi_performance}
\end{table}

\subsection{DP Rewriting}
\label{sec:rewriting}
Following the detection of IPI spans, each token in these spans is privatized using a modified version of the \textsc{DP-MLM} mechanism proposed by \citet{meisenbacher-etal-2024-dp} (licensed under MIT), which leverages Masked Language Models for DP text rewriting. In this way, a hybrid text privatization mechanism is created; as opposed to the original \textsc{DP-MLM}, which proposes the \textit{complete} rewriting of a text, \textsc{DP-IPI} solely rewrites identified IPI tokens, leaving all other text unchanged.

As per Algorithm \ref{alg:ipi}, tokens not detected as part of an IPI span are simply transferred to the output text. 
We note that as part of the modifications, we also implement an option to remove direct identifiers as detected by Microsoft Presidio\footnote{\scriptsize\url{https://microsoft.github.io/presidio/}} before IPI detection, in case the texts are not already de-identified, thereby adding an explicit anonymization step prior to DP rewriting of IPI spans.
We note that as part of the modifications, we improve the codebase to be able to handle arbitrarily long texts. 
This overcomes the previous limitations of \textsc{DP-MLM}, which is constrained to the 512 token context size of BERT-style encoder-only models. In particular, we implement a \say{sliding window} where the target token to be replaced is given the maximum-sized context during privatization. Details can be found in our code repository.

\begin{algorithm}[t]
\caption{\newline \textsc{DP-IPI} Private Text Rewriting Mechanism}
\label{alg:ipi}
    \begin{algorithmic}
        \small
        \Require input sentence $s  = w_1 \cdot w_2 \cdots w_n$, \\ per-token epsilon $\varepsilon$, \\ clipping values $C = (C_{min}, C_{max})$, \\ IPI-detected tokens $IPI$ \\ $REWRITE$ (e.g., DP-MLM, SANTEXT)
        \Ensure rewritten (privatized) document
        
        \State $\texttt{tokens} \gets tokenize(s)$ 
        \State $\texttt{private} \gets tokens$ 
        \For {$i \in 1...n$}
            \If {\texttt{tokens}[i] $\in IPI$}
                \State $\texttt{p} \gets REWRITE(\texttt{tokens}, \texttt{private}, i, \varepsilon, C)$
                \State $\texttt{private}[i] \gets p$
            \Else
                \State $\texttt{private}[i] \gets \texttt{tokens}[i]$
            \EndIf
        \EndFor
        \State \Return $detokenize(\texttt{private})$
    \end{algorithmic}
\end{algorithm}

In addition, we also apply the DP-IPI architecture to \textsc{SANTEXT} \cite{yue-etal-2021-differential}, a word-level DP mechanism that leverages constrained token sets for DP word replacements using the Exponential Mechanism. \textsc{SANTEXT} is well suited as a comparable method since it operates on the same level (word/token) as \textsc{DP-MLM}, allowing for the IPI modification. We also test the \textsc{SANTEXT+} variant, which only perturbs \say{sensitive} tokens based on corpus pre-training. We use the default parameters of this variant.
\paragraph{A note on $\varepsilon$ and privacy guarantees.}
As noted, an important consideration in any DP-based privatization is the mean of $\varepsilon$ (the privacy budget) with respect to the unit of privatization. 
The original \textsc{DP-MLM} by \citet{meisenbacher-etal-2024-dp} operates on the \textit{token-level}. Intuitively, this means that the unit of privatization, and the privacy guarantees resulting from this, are \textit{per token}. This also applies to \textsc{SANTEXT}, although we note here that \citet{yue-etal-2021-differential} introduce and use a \say{utility-optimized} metric DP notion, and as such, as choose $\varepsilon$ based on the original work.
In both cases, although the number of tokens being perturbed by \textsc{DP-IPI} will be inherently fewer; nevertheless, this updated mechanism and the original \textsc{DP-MLM} operate similarly regarding the privacy guarantees offered \textit{per-token}.
To mirror this in our privatization procedures and evaluation thereof, we select \textit{per-token} privacy budgets of $\varepsilon \in \{25,50,75,100\}$ for \textsc{DP-MLM} and $\varepsilon \in \{0.1,1,2,3\}$ for \textsc{SANTEXT}, and any token that is privatized (all tokens in \textsc{DP-MLM}/\textsc{SANTEXT} and IPI tokens in \textsc{DP-IPI}) are perturbed using these budgets. Both $\varepsilon$ ranges are chosen to mirror the sweeps tested in the original works. For a proof that \textsc{DP-MLM} and \textsc{SANTEXT} (and thereby, \textsc{DP-IPI}) are DP on the token-level, we refer the reader to the original works \cite{meisenbacher-etal-2024-dp,yue-etal-2021-differential}, respectively.
It is important to note that the DP guarantee solely for the \textit{perturbed} tokens in the \textsc{DP-IPI} variants.

\section{Experimental Setup}
We design two types of experiments: 1) utility experiments to assess the usability of the privatized data in different setups, and 2) empirical privacy experiments to report the risk of re-identification. 
The following introduces the data studied in this work and further specifications about the experiments. 

\subsection{Data}
\label{sec:data}

Following the repository\footnote{\scriptsize\url{https://github.com/bvanaken/clinical-outcome-prediction}} from \citet{van-aken-etal-2021-clinical}, we build a dataset of admission notes based on discharge summaries from the MIMIC-III to simulate patient state at admission. 

\subsection{Utility Experiments}

We benchmark the original version of \textsc{DP-MLM}, \textsc{SANTEXT}, and our \textsc{DP-IPI} mechanism across different $\epsilon$ values to assess the utility of the privatized admission notes in four downstream tasks: 
\begin{itemize}[leftmargin=0.15in]
    \itemsep -0.2em
    \item \textbf{Diagnosis Prediction} (Dia): a multi-label classification task (1,266 classes) in which the model predicts the clinical diagnoses in the form of ICD-9 codes based on patient state at admission.
    \item \textbf{Procedure Prediction} (Pro): a multi-label classification task (731 classes) in which the model predicts the ICD-9 codes of procedures (diagnostics or treatments) performed during the stay.
    \item \textbf{In-Hospital Mortality Prediction} (MP): a binary classification task in which the model predicts if the patient will survive the current stay.  
    \item \textbf{Length of Stay} (LoS): multiclass classification of the duration of the patient stay in the hospital. The four classes are: \textit{\say{< 3 days,} \say{3 to 7 days,} \say{1 week to 2 weeks,}} and \textit{\say{> 2 weeks.}}
\end{itemize}
The utility experiments results are reported as the average of three runs with different random seeds.

\paragraph{Additional Utility Metrics.}
We also capture the semantic similarity and text coherence between the original and privatized texts. 
Semantic similarity (S) is measured as the average cosine similarity between the embeddings of the original and private text counterparts. 
We use the average score of three embedding models for stability: \textsc{jina-embeddings-v3} \cite{sturua2024jinaembeddingsv3multilingualembeddingstask}, \textsc{all-mpnet-base-v2} \cite{reimers-gurevych-2019-sentence}, and \textsc{gte-large} \cite{li2023generaltextembeddingsmultistage}.

Text coherence (C) is measured via perplexity, using a \textsc{GPT-2} model \cite{radford2019language}, following previous work which uses perplexity as a proxy for text naturalness in text privatization \cite{10.1145/3485447.3512232}. 
We measure the mean perplexity from the first 128 tokens in each text. 

\subsection{Empirical Privacy Experiments}

We conduct empirical privacy experiments to assess privacy-preserving capabilities of our method, as compared to full \textsc{DP-MLM} and \textsc{SANTEXT}.
The following describes the two methods we used to evaluate empirical privacy protections.

\subsubsection{Text Re-Identification (TRI)}

The Text Re-Identification (TRI) method by \citet{manzanaressalor-etal-tri-2024} evaluates the disclosure risk of anonymized documents through simulating a machine-based attack. 
The premise is that the attacker has access to background information about a set of individuals which may or may not be included in the dataset. 
The attacker uses state-of-the-art neural language models to train a multi-class classification model which automatically links the background information to the individuals' documents.
Since our setup does not include access to other background information about the patients other than the information contained in the admission notes, we use a \textsc{Qwen2.5} model \cite{qwen2025qwen25technicalreport} to generate background information texts about each patient using the prompt shown in \Cref{fig:f_qwen_prompt} in \Cref{sec:qwen_prompt}. 
An example of a background information text generated from an admission note can be found in \Cref{fig:ex_generated_background} in \Cref{sec:ex_background}. 

We train the TRI model on the background information generated from the original documents. 
We then generate similar background information from the privatized versions of the original documents and assess the ability of the model to predict the right patients.
We conduct this experiment in two setups following \citet{manzanaressalor-etal-tri-2024}:

\paragraph{\textit{50\_eval} (50E).} This setup assumes a worst-case scenario for privacy and the easiest attack for an adversary, who has access to the background information of 50 individuals corresponding to all 50 documents in the dataset. 

\paragraph{\textit{500\_random} (500R).} A more practical scenario, where the attacker has access to background information from 500 people, but only 50 are present in the privatized dataset. 
Thus, the attacker model is trained to distinguish between 500 persons but is tested on its ability to identify the correct patient in 50 cases.
For evaluation reliability, we run each setup five times on different random test splits. 

\begin{table*}[ht!]
\centering
\resizebox{0.98\textwidth}{!}{
\begin{tabular}{lllllllllllll}
\hline
\textbf{$\epsilon$} & \textbf{Method}& \textbf{Dia} $\uparrow$ & \textbf{Pro} $\uparrow$ & \textbf{MP} $\uparrow$ & \textbf{LoS} $\uparrow$ & \textbf{S} $\uparrow$ & \textbf{C} $\downarrow$ & \textbf{500R} $\downarrow$ & \textbf{50E} $\downarrow$ & \textbf{LLM} $\downarrow$ & \textbf{$\gamma$} $\uparrow$ \\ \hline
- & De-identification (baseline) & 73.87 & 80.13 & 76.08& 66.75 & - & 90& 99.32\% & 99.60\% & 14.29 & -  \\ \hline
\multirow{7}{*}{25 / 0.1} 
 & DP-MLM (RoBERTa-base) & 65.48 & 66.92 & 66.83 & 61.19 & 0.80 & 1176 & 10.00\% & 26.80\% & 2.10 & 0.73 \\
 & DP-MLM (BioMed-RoBERTa-base) & 61.29 & 66.52 & 65.06 & 60.02 & 0.66 & 1972 & 1.28\% & 10.40\% & 1.35 & 0.76 \\
 & DP-IPI & 72.93 & \textbf{80.11} & \textbf{75.35} & \textbf{66.12} & \textbf{0.98} & \textbf{157} & 8.00\% & 26.00\% & 2.95 & \textbf{0.80}\\
 & SANTEXT & 50.76 & 56.34 & 52.06 & 52.66 & 0.44 & 2013 & \textbf{0.20\%} & \textbf{4.00\%} & \underline{\textbf{0.01}} & 0.71 \\
 & SANTEXT IPI & 72.61 & 76.71 & 71.01 & 65.79 & 0.95 & 4292 & 16.32\% & 34.80\% & 1.65 & 0.79\\
 & SANTEXT+ & 59.03 & 63.70 & 64.28 & 56.67 & 0.53 & 1488 & 1.88\% & 7.20\% & 1.09 & 0.72\\
 & SANTEXT+ IPI & \textbf{73.55} & 76.65 & 74.30 & 66.03 & 0.97 & 3262 & 21.72\% & 37.60\% & 2.11 & 0.76\\ \hline
\multirow{7}{*}{50 / 1} 
 & DP-MLM (RoBERTa-base) & 69.84 & 72.21 & 71.72 & 63.05 & 0.85 & 815 & 13.60\% & 26.80\% & 4.76 &  0.66\\
 & DP-MLM (BioMed-RoBERTa-base) & 68.68 & 70.55 & 68.57 & 61.24 & 0.76 & 1210 & 1.64\% & 8.00\% & 3.65 & 0.74\\
 & DP-IPI & \underline{\textbf{74.00}} & \textbf{80.53} & \textbf{75.13} & \textbf{66.52} & \underline{\textbf{0.99}} & \textbf{139} & 9.20\% & 26.40\% & 2.57 & \underline{\textbf{0.82}} \\
 & SANTEXT & 50.58 & 56.65 & 52.88 & 51.96 & 0.45 & 2664 & \underline{\textbf{0.04\%}} & \underline{\textbf{2.80\%}} & \underline{\textbf{0.01}} & 0.71 \\
 & SANTEXT IPI & 71.78 & 78.6 & 72.60 & 65.65 & 0.95 & 6496 & 17.68\% & 26.80\% & 1.72 & 0.80\\
 & SANTEXT+ & 62.85 & 63.77 & 65.99 & 57.64 & 0.53 & 1592 & 2.28\% & 9.20\% & 1.41 & 0.71\\
 & SANTEXT+ IPI & 72.45 & 79.98 & 73.09 & 65.77 & 0.97 & 3631 & 20.96\% & 39.60\% & 2.19 & 0.75\\ \hline
\multirow{7}{*}{75 / 2} 
 & DP-MLM (RoBERTa-base) & 71.07 & 74.36 & 73.00 & 63.43 & 0.86 & 709 & 15.20\% & 26.40\% & 5.01 & 0.66\\
 & DP-MLM (BioMed-RoBERTa-base) & 68.47 & 70.99 & 68.79 & 62.84 & 0.77 & 1106 & 1.84\% & 8.80\% & 4.19 & 0.72 \\
 & DP-IPI & \textbf{73.42} & \underline{\textbf{81.06}} & \underline{\textbf{75.81}}& \textbf{66.48} & \underline{\textbf{0.99}} & \textbf{136}& 14.00\% & 27.60\% & 2.78 & \textbf{0.79} \\
 & SANTEXT & 50.55 & 54.92 & 54.05 & 52.53 & 0.45 & 3850 & \textbf{0.20\%} & \textbf{3.20\%} & \textbf{0.03} & 0.71 \\
 & SANTEXT IPI & 71.36 & 76.19 & 71.78 & 66.09 & 0.95 & 10343 & 17.36\% & 33.60\% & 1.69 & 0.78 \\
 & SANTEXT+ & 62.98 & 64.48 & 62.85 & 57.41 & 0.55 & 1661 & 3.08\% & 10.00\% & 1.74 & 0.70 \\
 & SANTEXT+ IPI & 68.78 & 78.33 & 74.99 & 66.15 & 0.97 & 4132 & 21.00\% & 38.40\% & 2.59 & 0.74 \\ \hline
\multirow{7}{*}{100 / 3} 
 & DP-MLM (RoBERTa-base) & 71.11 & 76.43 & 73.21 & 63.86 & 0.86 & 678 & 16.00\% & 28.80\% & 4.78 &  0.66\\
 & DP-MLM (BioMed-RoBERTa-base) & 68.58 & 72.89 & 69.30 & 62.38 & 0.78 & 1069 & 2.08\% & 7.60\% & 3.99 & 0.74\\
 & DP-IPI & \textbf{73.31} & \textbf{80.43} & \textbf{75.74} & \underline{\textbf{66.82}} & \underline{\textbf{0.99}} & \underline{\textbf{135}}& 12.80\% & 26.80\% & 2.98 & \textbf{0.79} \\
 & SANTEXT & 51.33 & 56.20 & 53.14 & 53.43 & 0.47 & 5307 & \textbf{0.40\%} & \textbf{4.40\%} & \textbf{0.41} & 0.70 \\
 & SANTEXT IPI & 71.31 & 77.17 & 72.83 & 65.90 & 0.95 & 16027 & 19.00\% & 34.00\% & 1.97 & 0.77 \\
 & SANTEXT+ & 64.36 & 67.26 & 66.12 & 58.13 & 0.57 & 1652 & 4.44\% & 11.20\% & 1.83 & 0.71 \\
 & SANTEXT+ IPI & 72.99 & 78.72 & 73.41 & 65.85 & 0.97 & 4097 & 24.08\% & 43.60\% & 2.41 & 0.73 \\ \hline
\end{tabular}
}
\caption{Full experiment results for MIMIC-III. Scores in \textbf{bold} represent the best scores between all methods for each comparable $\varepsilon$, and the \underline{underlined} values denote the best scores per metric across all methods and $\varepsilon$ budgets. For $\varepsilon$, the left value is that used for DP-MLM (and variants), and the right is that used for SANTEXT (and variants).}
\label{tab:results}
\end{table*}

\subsubsection{LLM-as-a-Judge}
To support the TRI evaluation, we leverage LLM-as-a-Judge \cite{10.5555/3666122.3668142} for automatic evaluation of the degree to which personal information is protected in text privatization, in order to place a greater focus on \textit{personal identifier protection} as opposed to the re-identification risk focus of TRI.
In particular, we are interested in measuring the distinct \textit{number of personal identifiers} existing in the original text that are \say{leaked} to the private text. 
An effective privatization outcome would either mask or transform all such identifiers, while less effective privatization would simply transfer or insufficiently rewrite the spans. 
Thus, we craft a prompt that instructs the LLM to return the number of such leakages, where 0 is a perfect score and higher values represent worse privatization. 
We call this metric \textit{LLM}. 
This score is averaged across 100 randomly sampled original-private text pairs for each private dataset, using the same seed to ensure the same indices are selected across datasets. 
The full prompt can be found in Figure \ref{fig:llmj_prompt} of the Appendix. 
We call a \textsc{gpt-4o-mini} model, using the Microsoft Azure OpenAI service.

\subsection{Privacy-Utility Trade-off}
We calculate a privacy-utility trade-off ($\gamma$) which weighs relative gains in privacy against losses in utility, as compared to the non-private baselines \cite{mattern-etal-2022-limits,meisenbacher-etal-2024-dp}. 
We define the $\gamma$ score to be $\big(\frac{1}{2}\sum_i 1 - P_i^p / P_i^b\big) - \big(\frac{1}{3}\sum_j 1 - U_j^p / U_j^b\big)$, for $i \in \{TRI, LLM\}$ and $j \in \{Util, S, C\}$, where $b$ denotes baseline and $p$ denotes privatized. 
A positive score, therefore, implies that privacy gains outweigh utility losses, and vice versa. 
Note that \textit{Util} represents the average downstream performance score across all tasks. 
For coherence (C), we calculate a score relative to the original, non-private data, i.e., $C = C_b / C_p$ (intuitively, how close in perplexity the private data is to the original, with a max score of 1). 

\section{Results}

\paragraph{Utility.}

\Cref{tab:results} illustrates the utility results on the privatized admission notes. 
As expected, the utility drop is higher the lower the $\epsilon$ budget.
This is clearer for full-text rewriting using the original DP-MLM and SANTEXT implementations. Interestingly, using the vanilla \textsc{RoBERTa-base} model for \textsc{DP-MLM} achieves better utility across the board than the biomedical fine-tuned variant, indicating the benefit of generalized models for text rewriting.
The variance in utility is much lower when rewriting only detected IPIs using different $\epsilon$ budgets.
Specifically, rewriting only IPIs yielded similar utility to the original de-identified documents.
Importantly, \textsc{DP-IPI} achieved significantly higher utility in comparison to the original \textsc{DP-MLM} and \textsc{SANTEXT} implementations $(p < 0.05)$ in all downstream tasks and for all $\epsilon$ budgets tested.

\begin{figure}[t!]
\centering
\includegraphics[width=0.97\columnwidth]{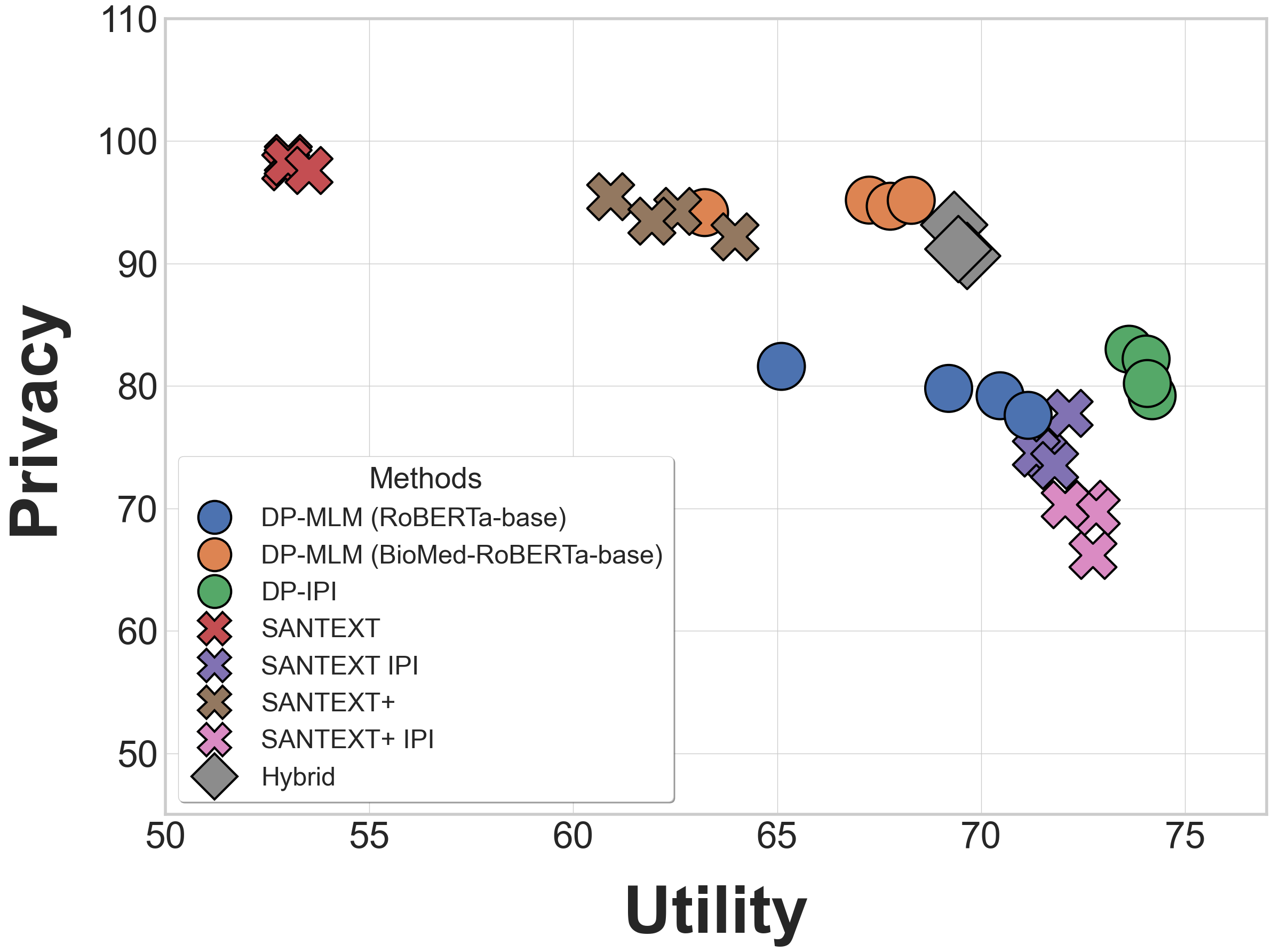}
\caption{Privacy-Utility trade-off Pareto frontier, including all methods tested in the main experiments and the hybrid setup, for each of the four tested $\varepsilon$ budgets. Utility is calculated as the average ROC-AUC ($\uparrow$) across the four downstream tasks in MIMIC-III. Privacy is calculated as $100 \ -$ average re-identification accuracy across attack scenarios (\textit{50\_eval} and \textit{500\_random}).}
\label{fig:utility}
\end{figure}

The benefits of \textsc{DP-IPI} are clear with respect to higher semantic reminiscence to the original texts, while being significantly more coherent and natural (perplexity). One can largely attribute these utility boosts to the immensely lower amount of tokens that are perturbed, a distinct hallmark of \textsc{DP-IPI}.

\paragraph{TRI.} 
The results of the TRI Attack in \Cref{tab:results} show that DP-IPI achieves a similar empirical privacy protection to the original DP-MLM method in both TRI scenarios.
Interestingly, privatized texts with DP-MLM (BioMed-RoBERTa-base) yield lower re-identification accuracy than DP-MLM (RoBERTa Base). 
This may be attributed to a strong shift from the original meaning in the rewritten texts with DP-MLM (BioMed-RoBERTa-base), which can also be inferred from the lower performance in the utility experiments.
Moreover, SANTEXT shows the highest robustness against the TRI attacks across all $\varepsilon$ budgets, potentially due to strong rewriting.
On the other hand, SANTEXT IPI fails to compete with its SANTEXT baselines.

\begin{table*}[ht!]
\centering
\small
\begin{tabular}{llcccccccccc}
\hline
\textbf{$\epsilon$} & \textbf{IPI Model}& \textbf{Dia} $\uparrow$ & \textbf{Pro} $\uparrow$ & \textbf{MP} $\uparrow$ & \textbf{LoS} $\uparrow$ & \textbf{S} $\uparrow$ & \textbf{C} $\downarrow$ & \textbf{500R} $\downarrow$ & \textbf{50E} $\downarrow$ & \textbf{LLM} $\downarrow$ & \textbf{$\gamma$} $\uparrow$ \\ \\ \hline
 \multirow{4}{*}{25} 
  &Baseline & 72.93 & 80.11 & 75.35 & 66.12 & 0.98 & 157 & 8.00\% & 26.00\% & 2.95 & 0.80 \\
   &BioMed-RoBERTa-base & 73.95 & 80.35 & 75.74 & 65.83 & 0.98 & 155 & 9.72\% & 23.60\% & 2.95 & 0.81 \\
   &DeBERTa-v3-large & 74.17 & 80.56 & 76.67 & 66.17 & 0.99 & 114 & 21.20\% & \underline{33.20\%}& 1.97 & 0.79 \\
   &BERT-Base & 73.64 & 79.75 & 74.30 & 65.92 & 0.98 & 157 & 8.16\% & 21.60\% & 2.99 & 0.81 \\ \hline
 \multirow{4}{*}{75} 
  &Baseline & 73.42 & 81.06 & 75.81 & 66.48 & 0.99 & 136 & 14.00\% & 27.60\% & 2.78 & 0.79 \\
   &BioMed-RoBERTa-base & 74.08 & 80.42 & 75.89 & 66.24 & 0.99 & 113 & 13.28\% & 33.20\% & 3.14 & 0.77  \\
   &DeBERTa-v3-large & 74.21 & 80.56 & 76.83 & 66.47 & 0.99 & 113 & 21.80\% & \underline{36.00\%}& 1.94 & 0.78 \\
   &BERT-Base & 73.97 & 80.15 & 75.18 & 66.03 & 0.99 & 137 & 12.76\% & 25.60\% & 2.98 & 0.79 \\ \hline
\end{tabular}
\caption{Empirical utility and privacy results for the IPI model ablation. Underlined values for privacy denote statistically significant difference from the baseline (one-sample $t$-test, $p < 0.05$) across five runs.}
\label{tab:ab1_results}
\end{table*}


\paragraph{LLM-as-a-Judge.}
In general, the \textit{LLM} results show less fluctuation across the different setups than the TRI attack scenarios, possibly because TRI uses medical or other contextual information to re-identify individuals when rewriting IPIs only. 
The high privacy protection (of personal identifiers) with \textsc{DP-IPI} in comparison to \textsc{DP-MLM} shows the effectiveness of using IPI detection on top of classical de-identification to omit personal information without rewriting all tokens. 
In the \textsc{SANTEXT} settings, \textsc{DP-IPI} offers similar protections, which however cannot match the nearly complete protection achieved by full \textsc{SANTEXT} rewriting.

\paragraph{Privacy-Utility Trade-offs.} Together, the results lead to higher privacy-utility trade-offs across all four privacy budgets, demonstrating the concrete value of \textsc{DP-IPI} on in-domain samples. This is not only demonstrated in Table \ref{tab:results}, but also in Figure \ref{fig:utility}, which demonstrates how \textsc{DP-IPI} achieves optimal trade-offs between privacy and utility. This is complemented by a hybrid, \textit{rise-aware} variant, discussed in the following section.

\section{Ablation Study}

To test the influence of the underlying IPI detection model, we study the privacy and utility of outputs produced different fine-tuned IPI detection models. 
We compare \textsc{DeBERTa-v3-large} \cite{he2021debertadecodingenhancedbertdisentangled}, \textsc{BioMed-RoBERTa-base} \cite{gururangan-etal-2020-dont}, and \textsc{BERT-Base} \cite{devlin-etal-2019-bert} to our baseline \textsc{RoBERTa-large}. 
The results of this study are found in \Cref{tab:ab1_results}. All models perform similarly on the held-out test after performing a hyperparameter search and choosing the best model for each setup (see \Cref{tab:ipi_performance_ablation}).
Despite some variation in the empirical privacy results, only \textsc{DeBERTa-v3-large} has a statistically significant difference from the baseline. 
These results, together with the privacy-utility trade-offs, suggest that the exact base model for IPI detection plays a less significant role for both utility and privacy.

\begin{table}[t]
\centering
\small
\begin{tabular}{@{}lccc@{}}
\toprule
\textbf{Model}        & \textbf{P} & \textbf{R} & \textbf{$\mathbf{F_1}$} \\ \midrule
RoBERTa-large (baseline)        & 0.81                & 0.94             & 0.87         \\
BERT-Base & 0.78& 0.92& 0.85         \\
DeBERTa-v3-large               & 0.83                & 0.94             & 0.88         \\
BioMed-RoBERTa-base & 0.83                & 0.93             & 0.88         \\
\bottomrule
\end{tabular}
\caption{Ablation results on the test set for \textsc{RoBERTa-large}, \textsc{DeBERTa-v3-large}, \textsc{BioMed-RoBERTa-base} and \textsc{BERT-Base} in \textbf{P}recision, \textbf{R}ecall, and $\mathbf{F_1}$. Support shows the number of examples in the test set.}
\label{tab:ipi_performance_ablation}
\end{table}

\begin{table*}[ht!]
\centering
\small
\begin{tabular}{llcccccccccc}
\hline
\textbf{$\epsilon$} & \textbf{Method}& \textbf{Dia} $\uparrow$ & \textbf{Pro} $\uparrow$ & \textbf{MP} $\uparrow$ & \textbf{LoS} $\uparrow$ & \textbf{S} $\uparrow$ & \textbf{C} $\downarrow$ & \textbf{500R} $\downarrow$ & \textbf{50E} $\downarrow$ & \textbf{LLM} $\downarrow$ & \textbf{$\gamma$} $\uparrow$ \\ \hline
 \multirow{2}{*}{25} 
  & DP-MLM & 65.48 & 66.92 & 66.83 & 61.19 & 0.80 & 1176 & 10.00\% & 26.80\% & 2.10 & 0.73 \\
  &DP-IPI& 72.93 & 80.11 & 75.35 & 66.12 & 0.98 & 157 & 8.00\% & 26.00\% & 2.95 & 0.80 \\
   &Hybrid& 69.76 & 74.20 & 70.74 & 62.66 & 0.84 & 609 & 2.48\% & 11.20\% & 4.79 & 0.72 \\ \hline
 \multirow{2}{*}{50} 
  & DP-MLM & 69.84 & 72.21 & 71.72 & 63.05 & 0.85 & 815 & 13.60\% & 26.80\% & 4.76 &  0.66\\
  &DP-IPI& 74.00 & 80.53 & 75.13 & 66.52 & 0.99 & 139 & 9.20\% & 26.40\% & 2.57 & 0.82 \\
   &Hybrid& 70.03 & 75.30 & 70.70 & 62.60 & 0.85 & 540 & 3.96\% & 14.80\% & 4.83 & 0.71 \\ \hline
 \multirow{2}{*}{75} 
  & DP-MLM & 71.07 & 74.36 & 73.00 & 63.43 & 0.86 & 709 & 15.20\% & 26.40\% & 5.01 & 0.66\\
  &DP-IPI& 73.42 & 81.06 & 75.81 & 66.48 & 0.99 & 136 & 14.00\% & 27.60\% & 2.78 & 0.79 \\
   &Hybrid& 69.89 & 73.67 & 71.49 & 62.72 & 0.85 & 531 & 4.84\% & 12.80\% & 4.74 & 0.72 \\ \hline
\end{tabular}
\caption{A comparison of DP-MLM (RoBERTa-base), DP-IPI, and  \textit{risk-aware} noise allocation, in which we apply lenient DP perturbation with $\varepsilon = 100$ on all non-IPI tokens to provide a minimal privacy guarantee on all tokens.}
\label{tab:hybrid_results}
\end{table*}

Furthermore, due to the imperfection of the IPI models, some IPI information might remain undetected. 
Therefore, we test a \textit{risk-aware} noise allocation setup, in which we apply lenient DP perturbation on all non-IPI tokens to provide a minimal privacy guarantee on all tokens. Here, we choose $\varepsilon = 100$. 
The results in \Cref{tab:hybrid_results} show that adding that this \say{light} privatization to non-IPI tokens provides further empirical privacy protection, but at the cost of further performance loss in utility. The benefit is, though, that despite rewriting all tokens, rewriting in this hybrid, risk-aware manner preserves data utility to a higher degree in comparison to rewriting the full text using lower $\varepsilon$. Above all, this ablation solidifies the notion that not all tokens should be privatized equally in DP text rewriting, making the case for hybrid schemes.

\section{Discussion}
\label{sec:discuss}

\paragraph{IPIs as a key to privacy protection.}
The empirical privacy results showcase an important lesson in text privatization and its evaluation. 
Modeling adversarial TRI, we demonstrate that even with \textit{significantly} fewer token perturbations, similar empirical privacy protections can be achieved. 
This indicates that it arguably makes very little sense to DP-rewrite all tokens when it is sufficient to privatize only identifiers within text. This becomes especially true in clinical texts, where the privatization of non-sensitive spans may serve to introduce new risks or potentials for misinterpretation. 
This, therefore, not only has profound implications for \say{medical utility} preservation, as can be drawn from Table \ref{tab:results}, but also emphasizes the need for selective and not \say{one-size-fits-all} privatization.

The design of our hybrid \textsc{DP-IPI} method also presents an efficiency gain in terms of computation time. Running full \textsc{DP-MLM} on a random sample of 100 texts from MIMIC takes 73.06 seconds (with the batched version), yet running the same setup with \textsc{DP-IPI} reduces the time needed to 48.88 seconds—a roughly 43\% reduction. This can be attributed to the far fewer perturbations needed (for example, 19 as opposed to 94 perturbed tokens in the example of Figure \ref{fig:dp-ipi}), despite the one-time IPI detection step (i.e. sequence classification). Along with solid privacy protections and significantly higher text quality, we argue that these efficiency gains make a strong case for the adoption of hybrid methods like \textsc{DP-IPI}.

\paragraph{Beyond metrics: Why a hybrid method makes sense.}
We extrapolate our analysis beyond the reported metrics of Table \ref{tab:results} to a qualitative critique of \say{blind} DP-based text privatization. 
Though the \textsc{DP-MLM} utility results may still be \say{acceptable} in certain cases, (i.e., utility is somewhat \say{preserved}), we refer the reader to the example provided in Figure \ref{fig:dp-ipi}. 
Performing such heavy and non-IPI-based privatization results in significantly perturbed texts, which may not only be semantically distant and incoherent, but may also be simply medically incorrect. 
In domains such as medicine, this may be non-negotiable, and our simple yet effective focus on IPIs provides a potential way forward for practical, usable DP-based text privatization. 
Thus our results support the hypothesis that re-identification attacks can be prevented by rewriting indirect non-medical information without necessarily rewriting the complete texts, the latter of which presents clear dangers beyond privacy.

These dangers are made concrete in the example of Figure \ref{fig:dp-ipi}, where key medical aspects in the text are perturbed significantly. For example, as shown in Figure \ref{fig:dp-ipi}, \textsc{DP-MLM} replaced the word \say{methadone} with \say{meth} and \say{54-year-old} with \say{young,} shifting the original meaning drastically. 
Not only are these perturbations sometimes unnecessary (with regards to personal identifiers), but they alter the meaning (e.g., medical exam/diagnosis) of the text beyond the point of factuality or usability.
Thus, full rewriting methods like SANTEXT inevitably suffer high utility losses and introduce erroneous information. 
\textsc{DP-IPI} addresses this issue by applying DP where personal identifiers pose a privacy vulnerability, but leaving semantics untouched in the absence of such risks.

Our focus on the clinical domain highlights a key challenge in DP text rewriting, a class of methods that is fundamentally grounded in the noisification of data to achieve privacy. We demonstrate, both using the example in Figure \ref{fig:dp-ipi} and empirically, that the effects of DP text privatization are severe and arguably unacceptable for high-stakes domains such as clinical notes. We propose, in turn, that selective and risk-focused privatization under DP must be \textit{hybrid}, and that favorable trade-offs can still be achieved at comparable privacy levels.

\paragraph{Lessons learned and limitations.}
In the design and evaluation of \textsc{DP-IPI}, we learn several lessons that lead to important implications for text privatization and its evaluation. In answer to domain-specific challenges as discussed above, an emphasis on IPIs may offer a potential solution, but this in turn places a dependency on the performance of IPI detection. Thus, it may be important for future work to annotate IPI information across languages and domains, clinical texts included, extending the work of research like \citet{baroud-etal-2025-beyond, baroud-etal-2026-multigrascco} and thereby creating more robust IPI extraction models as the foundation for DP text privatization.

In addition to greater annotation efforts, we argue that more work should be performed to study the potential side effects of DP text privatization, particularly those that may not be acceptable in medical environments. Here, paradigms such as \textsc{DP-IPI} might not only be more acceptable in terms of trade-offs (as we have demonstrated), but also crucial to maintaining the medical integrity of clinical notes, for example. For this, we suggest that human-centered study, especially with domain experts, are necessary to assess the practical acceptability of privatization schemes.

Finally, our experiments also shed light on the critical decision to be made in terms of \textit{empirical privacy} evaluation. 
Emphasizing the privatization of IPIs has proven highly effective in mitigating the attacker envisioned under TRI, whereas the framing of \say{authorship re-identification}, which models an attacker with full access to texts rather than \say{biographies}, is much more resistant. While a further critique of text privatization evaluation is outside the scope of this work, the diversity of available evaluation techniques, including LLM-as-a-Judge, calls for better systematization and standardization.

\section{Conclusion}
We introduce \textsc{DP-IPI}, a hybrid text privatization technique drawing on the strengths of Differential Privacy while focusing specifically on indirect personal identifiers in clinical texts. 
We highlight the strengths of DP-IPI, bolstered by in-domain IPI detection, in providing strong protections against re-identification while still reliably preserving downstream utility, semantic similarity, and coherence. 
Our work paves the way for hybrid DP/non-DP text privatization architectures, proposing an initial solution to the well-documented usability challenge. 
We share evidence of the feasibility of focused, efficient, and nevertheless effective DP-based text privatization, calling for further research on (1) practically usable, domain-specific DP text privatization, and (2) standardized evaluations thereof.

\section*{Limitations}
A primary limitation with our proposed method comes with its reliance on a fine-tuned IPI detection model, which is prone to lower performance for less frequent categories.
Future work should focus on creating a larger gold-standard dataset for detecting IPIs, either through annotating more documents or benefiting from synthetic data generation methods based on real examples. 

The evaluation of any Privacy-Enhancing Technology, especially when it comes to textual data, is a complex task that is still being actively studied. Thus, the results of the empirical privacy experiments are highly dependent on the methods used and the attack scenarios considered during the evaluation. Therefore, we adapted in our work various state-of-the-art methods and attack scenarios to present a comprehensive picture about the privacy-preserving capabilities of our method. In particular, we combine re-identification risk-based and LLM-based privacy evaluations, the latter of which focuses on the leakage of identifiers.

Specific to the integration of DP guarantees in text privatization with selective privatization, motivated from more traditional anonymization, our work sits directly at the intersection of two mindsets for text privatization. While we argue for and empirically demonstrate the benefit of such hybrid thinking, we also acknowledge its limitations. Firstly, selective DP privatization (i.e., leaving non-IPI tokens unperturbed for data release) can only be view comparatively (from a thereotical standpoint) at the \textit{token}-level. Similarly, comparability to non-DP based anonymization or de-identification methods is challenging, as these methods often redact or make use of placeholders, rather than perform \say{noisy} token replacement. We call for more work at the intersection of text privatization approaches, in order to bolster comparability and investigate novel ways to learn from different strengths. 

Finally, the effectiveness and generalizability of our hybrid method should be further tested in other domains, e.g., legal or social media texts, to test its applicability beyond the medical domain. 
For this, domain-specific gold-standard annotations are needed with special annotation guidelines developed by domain experts. We caution here that both detection performance, and as a result privatization effectiveness, can be highly dependent on both the quality of such annotations and the comprehensiveness of the annotation schemes. 

\section*{Ethical Considerations}
We confirm that the primary dataset utilized in this work, MIMIC-III, was used properly under its license of use, which included completing the necessary training to access the dataset.
Although the intended use MIMIC-III is not for patient re-identification, we do so for the purpose of evaluating empirical privacy protections, and not to optimize adversarial capabilities. 
Accordingly, in the experiments and ensuing evaluations, the utilized re-identification techniques were adopted from previous literature. 
MIMIC-III is a de-identified dataset, thereby ensuring no harm to real persons.

We acknowledge using AI assistance (ChatGPT and Google AI Studio) for generating the fictitious admission note shown in \Cref{fig:dp-ipi} and \Cref{fig:ex_generated_background} and for simple code generation for evaluation. 
We emphasized double-checking the generated code and content for validity and correctness.  

\section*{Acknowledgments}
We thank the anonymous reviewers for their time and feedback, which has helped shaped the final version of the work. We gratefully acknowledge funding from the German Federal Ministry of Research, Technology and Space (BMFTR) through the project VERANDA (16KIS2046K).

\bibliography{custom}

\appendix

\section{Reproducibility Notes}
\label{sec:rep_notes}
\paragraph{Model Training.}
The IPI detection model was created by fine-tuning a \textsc{FacebookAI/roberta-large} model \cite{DBLP:journals/corr/abs-1907-11692} for 15 epochs with early stopping.
All \textit{utility} models for the MIMIC-III dataset were fine-tuned for 150 epochs with early stopping patience of 10 epochs as in the original repository\footnote{\url{https://github.com/bvanaken/clinical-outcome-prediction/tree/master}} associated with the work by \citet{van-aken-etal-2021-clinical}. 
The Text Re-Identification (TRI) model was fine-tuned on 15 epochs using the configuration file from the original repository \footnote{\url{https://github.com/BenetManzanaresSalor/TextRe-Identification}}. 
The TRI training pipeline was configured to choose the model with the highest average accuracy across all privatization setups as the best model.

In all cases, model training was performed using all default parameters of the Hugging Face \textsc{Trainer} library, with a learning rate of 5e-5. Due to the context window limit of \textsc{BERT}-based models, all training described above was performed on texts truncated to a maximum of 512 tokens.

\paragraph{Randomization.}
For all procedures that involved randomization, a seed of 42 was used.

\paragraph{Infrastructure.}
For all procedures requiring GPU acceleration, including text privatization and model training, a single Nvidia RTX A6000 48GB GPU was utilized. 

\paragraph{Data.}
The MIMIC-III dataset\footnote{\url{https://physionet.org/content/mimiciii/1.4/}} is accessible following successful completion of an introductory course and data use agreement. It is licensed under the PhysioNet Credentialed Health Data License 1.5.0. In this work, we used a 50\% random sample of the full \textit{notes} dataset, which are de-identified free-text clinical notes.
Our final dataset for all four tasks includes 24426 unique texts, which are around 400 words long on average.

\paragraph{Text Rewriting.}
For all DP text rewriting, the reported $\varepsilon$ value indicates the per-token privacy budget used for any token perturbations. We note that all texts were truncated to be a maximum of 1024 tokens (as per the \textsc{RoBERTa-base} model of \textsc{DP-IPI}) prior to privatization.

\section{LLM Prompts for Evaluation}
\label{sec:qwen_prompt}
\Cref{fig:f_qwen_prompt} shows the prompt used to generate the background information based on the original or privatized admission note of each patient. 

Figure \ref{fig:llmj_prompt} provides the prompt used for the LLM-as-a-Judge privacy evaluation on both datasets.

\section{Example Generated Background Information}
\label{sec:ex_background}
\Cref{fig:ex_generated_background} presents an example of a generated background information generated from a fictitious admission note (prompt in \Cref{fig:f_qwen_prompt}).

\newpage

\begin{figure}[htbp]
\centering
\begin{tcolorbox}[width=0.97\linewidth, colback=orange!5!white, colframe=gray!75!black, title=LLM-as-a-Judge Evaluation Prompt]
\small
You will be given a user\_text and private\_text document.\\
Your task is to assess whether the private\_text contains any personal information or indirectly identifiable information from the user\_text.\\ \\
Give your answer as an integer, where 0 means that the private\_text is free of identifiers explicitly contained in the user\_text, and a positive integer means that the private\_text contains that many pieces of potentially sensitive or private information appearing in the original user\_text.\\

It is important ONLY to provide an integer as your rating, without any explanation.\\

Provide your feedback as follows:\\

Feedback:::\\
Rating: (your rating, as an integer)\\

You MUST provide values for `Rating:' in your answer.\\

Now here is the user\_text.\\

user\_text: [ORIGINAL\_TEXT]\\

private\_text: [PRIVATIZED\_TEXT]\\

Provide your feedback.\\
Feedback:::\\
Rating:

\end{tcolorbox}
\caption{The prompt used to produce the \textit{LLM} score for each of the original and private text counterparts.}
\label{fig:llmj_prompt}
\end{figure}

\begin{figure}[htbp]
\centering
\begin{tcolorbox}[width=0.95\linewidth, colback=orange!5!white, colframe=gray!75!black, title=Background Information Generation Prompt]
\small
--- SYSTEM PROMPT ---

You are an expert in data security and privacy. Your task is to extract social and personal details from clinical text to create a 'Background Knowledge' biography for a specific patient based on their admission note. This biography will be used to train an adversary model to assess the re-identification risk of the text after anonymization based on background knowledge.\\

--- USER PROMPT --- \\ \\
\#\#\# TASK:\\
Write a short (50-100 words) biographical summary about the following patient.
Use the **Clinical Note** below to extract 'Indirect Personal Identifiers' (IPIs) such as:\\
1. Occupation/Employment\\
2. Family\\
3. Habits (alcohol, smoking, drugs, sports)\\
4. Appearance such as height, weight, scars\\
5. Socio-economic or criminal history\\
6. Specific events before or during admission (accident, rejecting medication)\\

\#\#\# RULES:\\
1. Focus on the *narrative* (job, family, hobbies, the accident/event).\\
2. Do NOT include generic medical vitals (BP, HR) or medical information about the patient.\\
3. Focus on information that might be identifying about the patient.\\
4. Note that the texts are de-identified, that means that names and addresses are removed and dates are shifted, so they are not real values and should not be mentioned in the biography.  \\

\#\#\# CLINICAL NOTE:\\

\{MIMIC\_ADMISSION\_NOTE\}\\

\#\#\# OUTPUT FORMAT:\\ \\
\detokenize{[BIOGRAPHY_START]}\\ \\
The patient is a... (continue biography)\\ \\
\detokenize{[BIOGRAPHY_END]}

\end{tcolorbox}
\caption{The prompt used to generate background information for each patient based on their admission note.}
\label{fig:f_qwen_prompt}
\end{figure}

\begin{figure*}[htbp]
    \centering
    \begin{tcbitemize}[raster columns=1, raster row skip=0.5cm]
        
        \tcbitem[title=Original Admission Note,colback=orange!5!white, colframe=gray!75!black]
        \small
        SERVICE: MEDICINE \\

HISTORY OF PRESENT ILLNESS: \\
Mr. [Known patient lastname] is a 54-year-old male with a history of hypertension and substance use disorder who presents to the ER with shortness of breath and palpitations. The patient reports that symptoms began "right after the Thanksgiving Day parade ended" while he was watching it from his porch. He describes a sudden onset of chest tightness, rated 7/10, non-radiating. He initially attributed it to stress regarding his upcoming court date but the symptoms persisted.\\

PAST MEDICAL HISTORY:\\
Hypertension.
Chronic obstructive pulmonary disease (COPD).
History of opiate abuse, currently on methadone maintenance.
Hepatitis C (untreated).
Traumatic brain injury (sustained during a prison riot in [2138]).\\

SOCIAL HISTORY:\\
Family: Patient lives with his wife who is currently working as the head librarian at the local community college. He has two estranged daughters; one is reportedly a prosecutor in [State] which the patient cites as a major source of family tension.
Socioeconomic: The patient was formerly a union carpenter but has been unemployed for 6 years following his felony conviction. They currently reside in a rent-controlled apartment in the [Neighborhood] district, near the old steel mill.
Criminal: Extensive legal history. Patient was incarcerated for 8 years for armed robbery and was released 4 years ago.
Substance Use: Tobacco 1ppd x 30 years. Alcohol: socially, admits to a "few beers" during the [University] football game last Saturday.\\

ALLERGIES:

Penicillins (Causes Hives). \\

MEDICATIONS ON ADMISSION:

Amlodipine 10 mg daily.

Methadone 80 mg daily (verified with clinic).

Albuterol Inhaler PRN.\\

PHYSICAL EXAM:

VS: T 98.4, HR 110, BP 155/92, RR 22, O2 sat 94\% on RA.

Gen: Anxious appearing male, diaphoretic, wearing an ankle monitor.

HEENT: PERRL, EOMI. Sclera anicteric. Poor dentition.

CV: Tachycardic, regular rhythm. No murmurs, rubs, or gallops.

Lungs: Diffuse expiratory wheezes bilaterally. Prolonged expiratory phase.

Abd: Soft, non-tender, non-distended. Tattoo of a spider web visible on the left flank.

Ext: No edema. 2+ radial pulses. Finger tattoos fading.
        
        \tcbitem[title=Generated Background Information,colback=orange!5!white, colframe=gray!75!black]
        \small
        The patient is a 54-year-old male who was formerly a union carpenter but has been unemployed for the past six years following his felony conviction. He lives with his wife, who works as the head librarian at the local community college, in a rent-controlled apartment near an old steel mill. The patient has two estranged daughters; one is reportedly a prosecutor, contributing to significant family tension. He was incarcerated for eight years for armed robbery and recently attended a [University] football game where he had a few beers. A tattoo of a spider web is visible on his left flank.
        
    \end{tcbitemize}
    \caption{An example of a background information text generated using the prompt presented in \Cref{fig:f_qwen_prompt} from a fictitious admission note similar to the ones in MIMIC-III.}
    \label{fig:ex_generated_background}
\end{figure*}

\end{document}